# Deep Phasor Networks: Connecting Conventional and Spiking Neural Networks

Wilkie Olin-Ammentorp[1,2], Maxim Bazhenov[1,2]

1 Department of Medicine, University of California, San Diego

2 Institute for Neural Computation, University of California, San Diego

*Corresponding Author: Maxim Bazhenov

**Email:** mbazhenov@health.ucsd.edu

**This PDF file includes:**

> Main Text
> Figures 1 to 8

**Abstract**

In this work, we extend standard neural networks by building upon an assumption that neuronal activations correspond to the angle of a complex number lying on the unit circle, or 'phasor.' Each layer in such a network produces new activations by taking a weighted superposition of the previous layer's phases and calculating the new phase value. This generalized architecture allows models to reach high accuracy and carries the singular advantage that mathematically equivalent versions of the network can be executed with or without regard to a temporal variable. Importantly, the value of a phase angle in the temporal domain can be sparsely represented by a periodically repeating series of delta functions or 'spikes'. We demonstrate the atemporal training of a phasor network on standard deep learning tasks and show that these networks can then be executed in either the traditional atemporal domain or spiking temporal domain with no conversion step needed. This provides a novel basis for constructing deep networks which operate via temporal, spike-based calculations suitable for neuromorphic computing hardware.

**Significance Statement**

This work demonstrates a novel technique of constructing and training deep neural networks which can then be accurately and efficiently evaluated via networks of resonate-and-fire neurons. This method provides a potential link between artificial and biological computation and has immediate applications for neuromorphic, oscillatory, and photonic computing methods. These networks may also be integrated with other areas of research, including vector-symbolic architectures and temporal backpropagation with surrogate and/or adjoint gradients.

**Main Text**

**Introduction**

Efficient learning and inference processes remain a challenge for deep-learning based artificial intelligence methods (1). Problems include poor generalization beyond properties of the training data (2), catastrophic forgetting during new task learning (3), and lack of transfer of knowledge (4). Many efforts focus on addressing this by carrying out deep learning via the use of networks which communicate via binary events - 'spikes,' bringing artificial neurons closer to their highly efficient biological counterparts (5–7). While recent efforts to link traditional deep neural networks (DNNs) to spiking neural networks (SNNs) via the usage of rate-coding have yielded SNNs which can attain high accuracy, this achievement comes with several caveats. Firstly, converting a suitable DNN to a rate-based SNN can require complex conversion methods to normalize weights and activation values (8–10). Secondly, the resulting spike-based networks lack many established characteristics of biological computation, such as fast inference and sensitivity to the timing of individual spikes (11–15). Lastly, even when executing on specialized neuromorphic hardware, these spiking networks require processing long sequences of spikes to evaluate rate and as a result provide at best a marginal gain in energy efficiency when compared to traditional networks running on conventional hardware (16). For these reasons, to create a spiking network which can achieve the goals of neuromorphic computing (such as high performance, energy efficiency, and biological relevance), alternate approaches from rate-coding will likely be required (17–21).

We build on previous works which assume that the state of a neuron may be represented by the angle of a complex number, commonly referred to as a "phasor" (22,23). Phasors are often used in electrical engineering and physics to provide convenient representations and manipulations of sinusoidal signals with a common frequency (24). A phasor describes a sinusoidal signal's phase relative to a reference signal, and can be scaled by a real magnitude to describe any complex number in polar form (Equation 1).

Equation 1.    $A\angle\varphi = Ae^{i(\omega t+\varphi)} = \cos(\omega t+\varphi) + i\sin(\omega t+\varphi)$

Any set of sinusoidal signals with a common frequency can be represented by a vector of phasors, and as their superposition produces another sinusoid, it can also be represented by a single new phasor. This value is calculated by summing the original vector of phasors in their complex form. The phase of the resulting complex number is calculated via a non-linear trigonometric operation, allowing the superposition of phasors to form the basis of a neuronal activation function suitable for a deep neural network. The representation of information through phasors and non-linear properties of their superpositions provides the basis of information processing within phasor networks. While the apparent phase of a signal varies with respect to time, either an absolute starting time or reference signal can be used to decode phases from a set of time-varying signals.

In this work we propose an extension of the "classic" deep neural network architecture by replacing real-valued activations by phasors. We describe the operations needed to calculate propagation of the activation through the phasor network and show that the phasor network can be executed either in an atemporal domain or temporal domain. In the former, values are passed between layers in standard tensors of phasor values. In the latter, values are passed between layers by series of precisely-timed binary spikes, referred to as 'spike trains'. Atemporal execution is well-suited to existing computer architectures (e.g. CPU and GPU), and temporal execution is suited to current or future neuromorphic platforms. We demonstrate that either execution mode leads to similar performance on standard machine learning datasets. We compare the temporal networks to alternatives employing different spike-code based methods and show that phasor networks offer a unique trade-off between efficiency and robustness.

## Results

### Atemporal Evaluation
*Activation Function*

Let us assume that the input to a single neuron consists of a vector of phases, $x$. For convenience and ease of integration within existing deep learning frameworks, in this work all phase angles are reported after being normalized by π, so that $x \in [-1,1]$. Multiplying these values by π converts to a standard angle in radians.

To compute a single neuron's output $y$ given an input vector $x$ with $n$ elements, the real valued phases $x$ are converted into an explicit complex representation. These complex elements are then scaled by a vector of corresponding weights $w$, which we currently restrict to be entirely real-valued. The sum of the scaled complex elements produces a new complex value with an amplitude and phase (Equation 2). To extract only its phase, the two-argument arctangent ($atan2$) is applied (Equation 3). Through these operations, a nonlinear neuronal operation is obtained. Additionally, the local continuity of these operations ensures deep networks employing them can be optimize each neuron's weights $w$ using standard backpropagation techniques.

Equation 2.    $x' = \sum_{i=1}^{n} e^{i\pi x_i}$
Equation 3.    $y = atan2[Im(x'), Re(x')]$

*Output and Loss Functions*

In an image classification task, the output of a standard network is often a vector of real values normalized between zero and one using the softmax function (25). The outputs of such a network are then taken to be the probability that the input image belongs to the corresponding output class, and a loss function such as cross-entropy is used to minimize the divergence between the network's predictions and the image's ground-truth class.

In contrast, in a phasor network the output layer creates a vector of phases. Given an input image with a label $c$ out of a total of $n_c$ classes, the network predicts this class at the output by causing the corresponding neuron in the output layer to produce a phase which is in quadrature (90° out-of-phase) to all others (Equation 4). A loss function using cosine similarity of the vector of predicted values ($\hat{y}$) to the target vector ($y$) is used to measure the convergence of the network's output to the desired phase encoding (Equation 5). For an input with an unknown label, the predicted class $\hat{c}$ is produced by finding the neuron which produces a phase closest to 90° (Equation 6).

Equation 4. $\quad y = \frac{1}{2} \cdot onehot(c, n_c)$

Equation 5. $\quad loss(y, \hat{y}) = 1 - \cos[\pi \cdot (y - \hat{y})]$

Equation 6. $\quad \hat{c} = argmin[abs(\hat{y} - \frac{1}{2})]$

*Image-to-Phase Conversion*

One issue in implementing phasor-based networks arises at the input layer of such a network. Inputs such as images are almost always encoded on a domain with pixel intensities normalized between 0 and 1. However, as previously stated, a phasor network utilizes inputs on the domain $[-1,1]$. We show below that an initial conversion step between domains can assist phasor networks in reaching performance levels that match conventional networks. A simple linear scaling ($2x - 1$) between domains is insufficient, as this will lead to ones and zeros being encoded into phases (-1 and 1) which have an identical cosine similarity. Instead, we utilize two intensity-to-phase conversion methods: the first is a 'normalized random projection' (NRP), and the second a 'random pixel phase' (RPP).

The NRP method constructs a sparse random projection by sampling values from a uniform distribution on the domain [-1,1]. The input image is then multiplied by this square matrix to produce a new input vector of the same dimension with pixel data distributed across multiple values. A simplified batch-normalization layer with two moving moments learned during training (mean and standard deviation) is applied to the random projection to keep approximately 99% of projected phases within the range [-1,1]. Lastly, outlier values are clipped to this domain. Alternatively, the RPP method randomly chooses input pixels which are either multiplied by 1 or -1. Collectively, this spreads values across the domain [-1,1] although the absolute range of an individual pixel does not increase.

*Model Architecture and Accuracy*

To show that deep phasor networks can be effectively trained using the approach described above, we train a series of standard and phasor-based image classifiers on the standard MNIST, FashionMNIST (F-MNIST), and CIFAR-10 datasets (26–28).

First, we demonstrate a simple multilayer-perceptron (MLP) model with one hidden layer. The architectures of the MLP networks trained on MNIST-format images (28x28x1 pixels) are identical, consisting of an input layer, intensity-to-phase conversion method, hidden layer of 100 neurons, and an output layer of 10 neurons, with biases are disabled for all neurons. A neuronal dropout rate of 25% is used for regularization (Figure 1a). To provide a control for comparison, 'standard' versions of this model are constructed by substituting a ReLU activation function at the hidden layer and softmax at the output. Experimental phasor models instead use the previously described activation method.

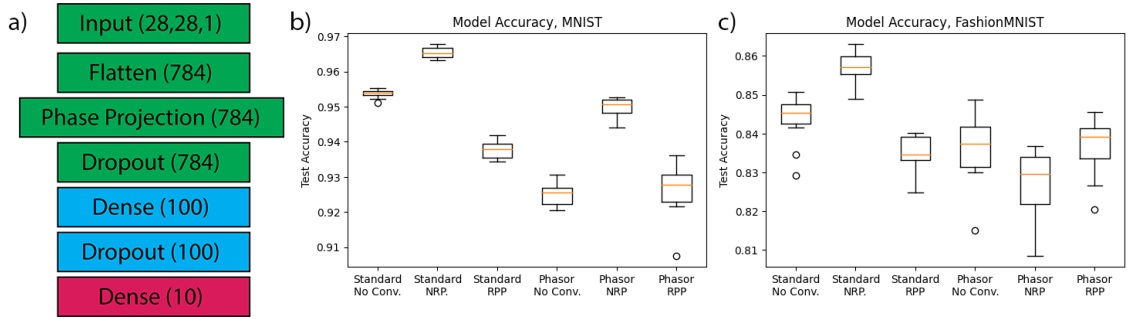

**Figure 1.**
Architecture of the multi-layer perceptron (MLP) network used. Each layer is labeled with its output shape. (b) Classification accuracy of this model on the standard MNIST dataset and (c) FashionMNIST dataset over 12 trials. 'Standard' control models use a ReLU/softmax activation function and phasor models use the phasor activation method described in the main text. NRP models apply a normalized random projection to convert intensities to phases and RPP models randomly select pixels to produce positive or negative phases. These results suggest that the phasor activation function can provide an effective basis for constructing deep networks.

In these results, we train groups of 12 models and report the group's mean accuracy plus or minus its standard deviation. Phase projection layers are included even in standard networks as they can affect test accuracy, particularly for the MNIST dataset.

With an NRP conversion, the standard models reached a test accuracy of 96.4±0.1% on standard MNIST after 2 epochs of training. Similarly, the phasor models reached a median accuracy of 95.0±0.3%. Training instead on F-MNIST dataset, the standard models with NRP conversion reached 85.7±0.4% accuracy on the test set (Figure 1b). The highest-performing phasor networks for F-MNIST used the RPP conversion, with an accuracy of 83.7±0.7% (Figure 1c).

To classify the CIFAR-10 dataset, a convolutional architecture is used. It consists of an input block, two convolutional blocks, and a dense output block (Figure 2a). The input block consists of an optional phase projection method followed by a batch-normalization step. The first convolutional block consists of two convolution layers with 32 channels and 3x3 kernels, followed by a 2x2 max-pool and 25% dropout. The second convolutional block is identical but uses convolutional layers with 64 channels. An additional L2 regularization is applied to kernels in all convolutional layers. The dense block flattens the convolutional output, and applies a dense layer with 1000 neurons, a dropout of 25%, and a final dense layer of 10 neurons. Biases are again disabled on all neurons. All layers in the 'standard' version of the model use a ReLU activation, except for the final dense layer which uses a softmax. No phase projection is used for the standard network, and RPP is used for the phasor network.

For the convolutional architecture, one additional change is made between control and phasor-based networks. In the latter case, a minimum-pool (min-pool) is substituted for the more common max-pool on the basis that in the temporal representation of a phase value, lower values correspond to earlier spikes. In biological networks, earlier spikes would result from stronger synaptic inputs and would more likely contribute to the later processing while later spikes may be ignored or canceled (29). Min-pooling also may be possible to approximate based on a combination of winner-take-all circuits and network dynamics (30). We find that the substitution of a min-pool operation for max-pool operation in the phasor CIFAR-10 networks leads to no significant change in performance. However, average pooling leads to significant performance loss and is not used.

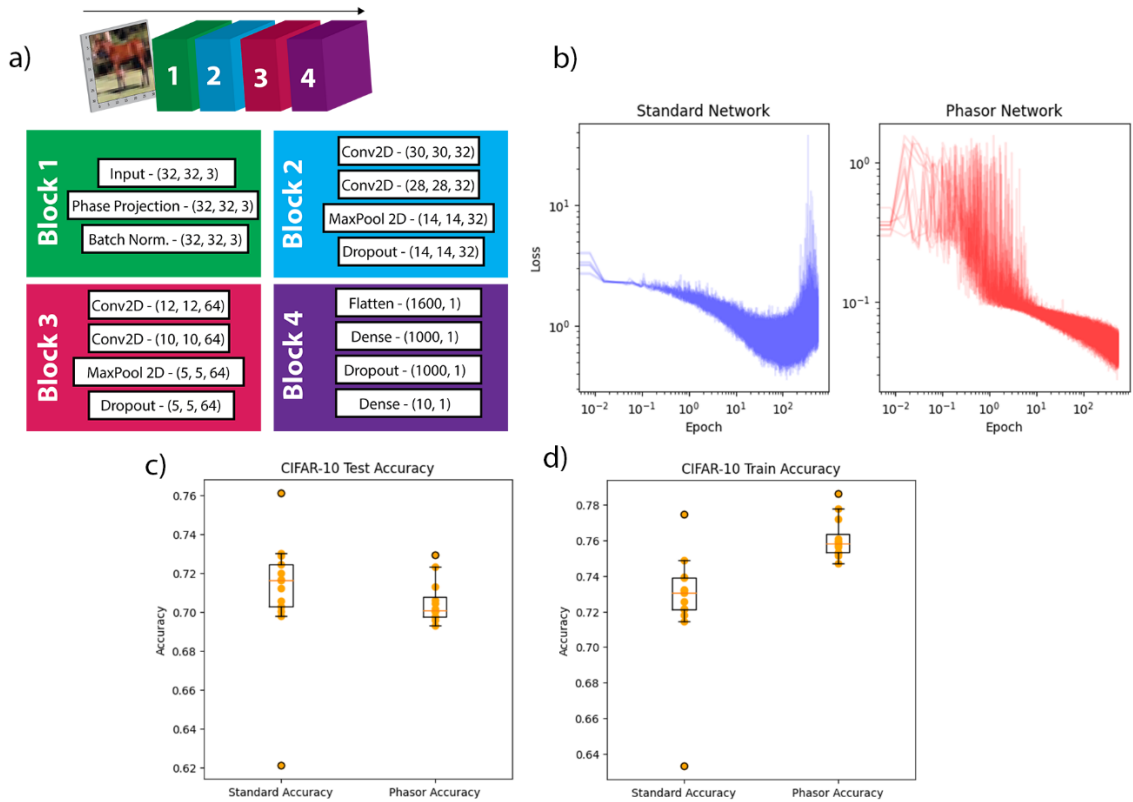

**Figure 2.**
(a) Architecture used for the CIFAR-10 image classification task. This 15-layer network utilizes operations required for modern image classifiers such as convolutional layers and spatial pooling. This architecture is implemented with ReLU activation functions (except for a final softmax) to produce a 'standard' control network. In phasor networks, all layers use the phasor activation method and a minimum-pool operation is substituted, but the architecture otherwise remains the same. (b) Loss curves during network training show a common pattern for both network types, though the loss of the standard networks in later epochs is more greatly affected by the included L2 loss on convolutional kernel weights. (c) Phasor and standard networks reach the same classification accuracy on the test set (no significant difference, p=0.58, n=12). (d) There is a slight but significant difference in the accuracy on the training set, with phasor networks producing significantly higher accuracy (p=0.001, n=12), suggesting they may require alternate regularization methods.

Both network types (standard and phasor) are trained using an augmented dataset to prevent overfitting. A rotation range of 15°, a width and height shift range of 10%, and 50% chance of a horizontal flip were used. All networks were trained using 70312 batches of 128 augmented images, corresponding to approximately 180 epochs on the original training dataset.

We found that in case of this more complex network architecture, no significant difference is observed between the standard and phasor networks' performance on the test set, with accuracies of 71.1±3.1% and 70.5±1.1% respectively (Figure 2c). This contrasts with performance differences observed in the MNIST and F-MNIST tasks, and may be due to the higher regularization penalties imposed on both networks and the more complex image recognition task. The kernel weight regularization may effectively penalize the standard network more than the phasor network with the same parameters, as it can be observed impacting standard network loss more in later epochs (Figure 2b) and the accuracy of the phasor networks on the original CIFAR-10 training set is significantly higher (Figure 2d).

To summarize, these results demonstrate that applying phasor-based representation and using the phasor neuron described in Equations 2-3 can create networks which achieve results on-par with standard neural networks. Some differences in performance between phasor and standard networks can be attributed to many factors, including the image-phase conversion method, regularization techniques, and the dataset. Next, we investigate an alternate inference mode for phasor networks which is inherently temporal in nature.

**Temporal Evaluation**

As shown in Equation 1, the invariants of a temporal sinusoidal signal can be used to represent it in a compact, atemporal form which can be easily manipulated. This equivalency allows for the convenient analysis of systems such as alternating-current (AC) electrical circuits. And similarly to the case of AC circuits, a neural network utilizing phasor-based activations also has an equivalent, temporal form which may be implemented by the dynamics of a virtual or potentially physical system. This equivalence was previously reported by Frady et al., who used it to execute an associative memory (31); here it is employed to execute deep neural networks.

In this section, we describe how one version of a temporal phasor model can be constructed. In contrast to the previous atemporal execution mode, phases are communicated between neurons not by tensors of floating-point values, but by periodic and precisely-timed binary 'spikes.' The integration of currents induced by these spikes excite the neuron to fire in response, allowing for an equivalent temporal calculation of what was previously computed solely via standard linear algebra. This equivalence allows for different hardware systems to carry out the computations required for a deep neural network.

*Spike-Based Representation of Phase*

In our model, the tensors of phase values communicated between neurons represent the relative phases between a sinusoidal signal and a reference. Previously, each value was represented by a single, real number restricted to the domain $[-1, 1]$. However, the relative phase can be represented in a different form: an instantaneous pulse or 'spike' can be used to mark whenever a sinusoid reaches its maximum (Figure 3a). This creates a sparse, periodic representation of the underlying signal which communicates the same phase information. For instance, a signal in-phase to a reference has an atemporal representation of 0. In the spiking, temporal domain, this is represented instead by a series of spikes which occur in the middle of the period defined by the reference signal. Signals with lower values will precede the cycle's midpoint, and signals with higher values will follow it (Figure 3d).

In this manner, the representation of a phase is changed from a single value to a series of instantaneous spikes. To compute the same activation function as was previously demonstrated, the phases represented by these signals must be weighted, superimposed, and the phase of the output signal calculated. Essentially, this is accomplished by each spike exciting a sinusoidal current within a resonate-and-fire (R&F) neuron which then accumulates through time into a voltage. The peak of this voltage is detected and used in turn to create an output spike representing the same activation value as in the atemporal calculation (Figure 3e). Below we show this operation in detail.

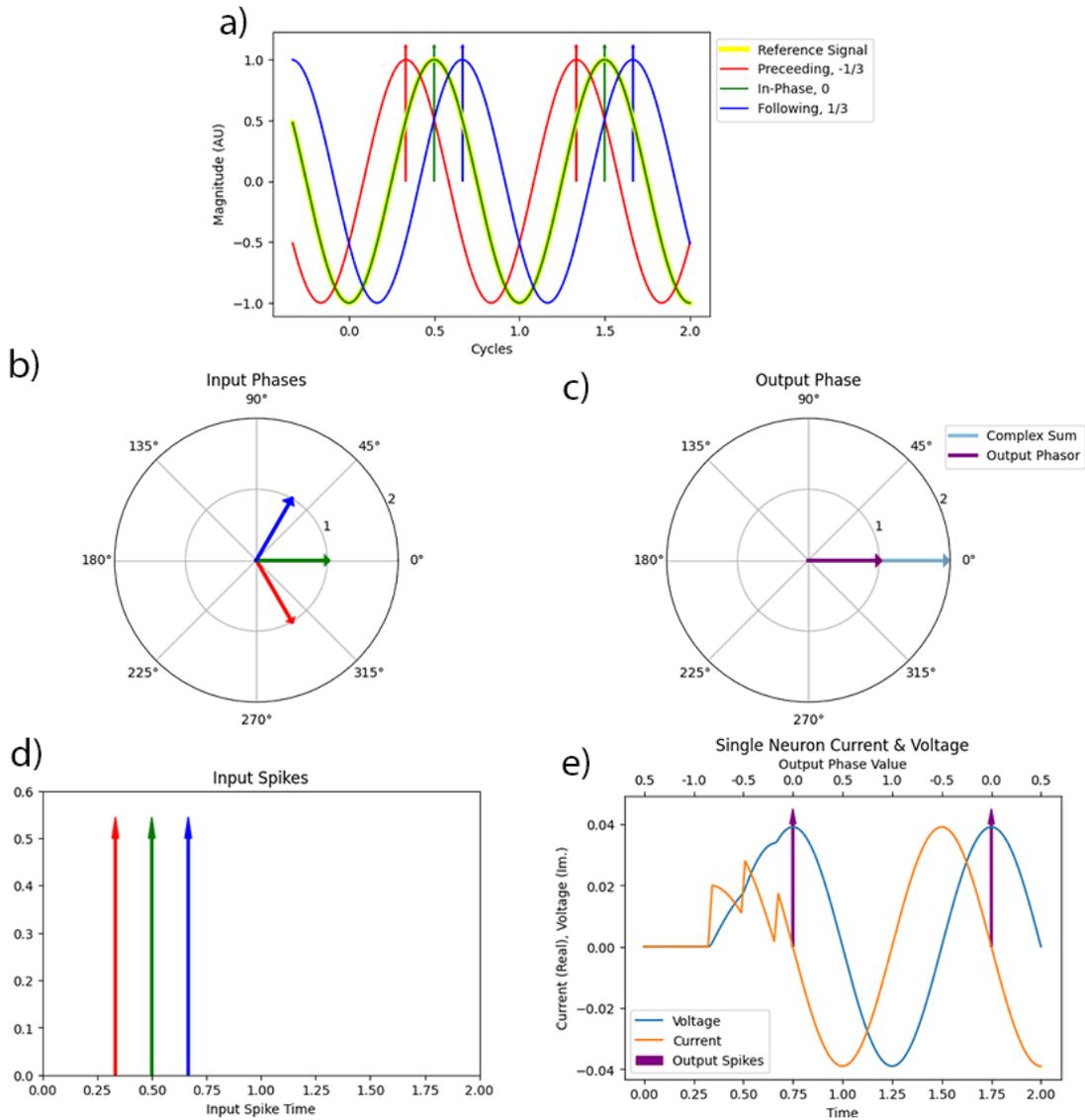

**Figure 3.**

Phasors capture the invariants of a time-varying sinusoidal signal (a) and allow them to be manipulated using complex arithmetic. For instance, summing three input phasors (b) produces a sum that has its own corresponding phase value (c). Alternately, this calculation can be carried out by representing the phase of the input signals with timed spikes (d) which excite sinusoidal currents in a neuron which are summed to produce a new output spike, representing the phase of the sum (e). These calculations are equivalent, but can utilize different forms of storing, transmitting, and processing information.

*Equivalent Resonate and Fire Model*

Taking a derivative of a complex number $z$ representing a harmonic signal with respect to time (Equation 1) gives Equation 8, which is identical to the case of an R&F neuron with no 'leakage' (attraction to the rest state) (32). This shows that the dynamics of the R&F neuron are inherently linked to the calculation which must be carried out in the temporal execution of a phasor network. Below, leakage is re-introduced to allow for inferences which change with respect to time. But in the case of an R&F neuron with no leakage, the voltage-current oscillation produced after a current pulse is identical to the rotation of a phasor with respect to time represented in Equation 1. This allows the R&F neuron to 'recreate' the sinusoidal signals represented by the series of input spikes. Here, following the original R&F convention we refer to the 'current' $U$ of an R&F neuron as the real part of its complex potential $z$, and its 'voltage' $V$ as the imaginary part of $z$ (32).

Equation 7.  $z = Ae^{i\omega t + \varphi}$

Equation 8.  $\frac{\partial z}{\partial t} = (0 + i\omega)z \rightarrow (-b + i\omega)z$

With this approach, the superposition and phase detection which was previously calculated atemporally using Equations 1-2 can be carried out exactly in the temporal domain using an R&F neuron with no leakage. First, the duration of one cycle of a reference signal is defined as $T$, giving the R&F neuron a matching natural angular frequency $\omega$ of $2\pi/T$. Phases of an input vector $\mathbf{x}$ are represented by the spikes - Dirac delta functions which occur once per each cycle $T$. Real-valued weights are represented again with $\mathbf{w}$ (Equation 9).

Equation 9.  $\frac{\partial z}{\partial t} = i\omega z + \sum_{i=1}^{n} \mathbf{w}_i \delta(t - \mathbf{x}_i)$

Equation 10.  $z(T) = \sum_{i=1}^{n} \mathbf{w}_i \cdot e^{-i\pi x_i} \quad if \ z(0) = 0$

Integrating Equation 9 through a single cycle (t=[0,2], if period $T = 2$) produces another form of the superposition of complex values (Equation 10, proof in methods). The timing of the inputs $x$ defines the phase of the output $z$, which can be determined by detecting a local peak in the voltage value. This allows the full calculation required for a phasor neuron (complex superposition & phase measurement) to be carried out in the temporal domain using input spikes.

However, leakage must be retained within R&F neurons if they are to carry out different computations through time (allowing their potentials to gradually return to the initial condition required in Equation 10). By increasing the level of leakage $b$ in the R&F neuron from 0, the same approximate calculation can be carried out, ideally without having a major effect on the phase of the superposition. Too high a leakage value will cause the 'memory' of the neuron to be too short, leaving it unable to calculate an approximation of Equation 10 as the resulting oscillations will decay too quickly to achieve superimposition. However, too low a leakage value will prevent the neuron from returning to a rest state which is required to allow it to adapt flexibly to new computations through time. To strike a balance between these extremes, we use a leakage set to 1/5th the value of the integration period $T$ (Table 1).

**Table 1.** Parameters used for temporal phasor neurons.

| Parameter | Value |
| --- | --- |
| Period ($T$) | 1.0 s |
| Leakage ($b$) | 0.2 |
| Box width scale ($s$) | 0.05 |
| Threshold ($V_{th}$) | 0.03 |
| Refractory Period | 0.25 s |

The representation of an input spike by an instantaneous Dirac delta function at a time $x$ can be relaxed by convolving it with a kernel such as a box function ($\Pi(t)$ with width scale factor $s$). These alterations (leakage and box kernel) are included in Equation 11, which is solved numerically through time to calculate the complex potential $z$ of a phasor neuron. To reiterate the other parameters in this equation, $T$ is the R&F neuron's fundamental period, $b$ is a positive value which sets the neuron's leakage, $w$ is the vector of the neuron's $n$ real-valued input weights, and $t$ is time. In Table 1 we present the values used for these parameters in our experiments.

Equation 11. $\quad \frac{\partial z}{\partial t} = \left(-b \cdot T + \frac{i2\pi}{T}\right) z + \sum_{i=1}^{n} w_i \Pi(s \cdot t - x_i)$

The spiking threshold of an R&F neuron is an important parameter. To meet the requirements of a phasor network, a spike is produced from the R&F neuron when it reaches a certain phase of its current/voltage oscillation. This phase can be found by determining when the neuron's complex potential sweeps through a defined set of conditions. Here we define a set of conditions which causes the R&F neuron to produce a spike: (a) its imaginary value (voltage, $V = Im(Z)$) reaches a maximum, i.e. $\partial Im(Z)/\partial t = 0$ and $Im(Z) > 0$; (b) its voltage is above a set threshold ($Im(Z) > V_{th}$), and (c) the time after last preceding spike exceeds a short refractory period ($T/4$). The last condition reduces the occurrence of multiple spikes in the same period. This gradient-based method allows the output spikes to be sparsely produced without referencing an external clock. For conciseness and given their approximate equivalence, we term an R&F neuron using this spike detection method as a temporal phasor neuron.

*Demonstration of Equivalent Neuron*

Next, we first demonstrate that a temporal phasor neuron can carry out the identity function. A series of phasor neurons are excited with single input spikes which repeat with a frequency $1/T$. The relative timing of these spikes communicates the phase of the input they represent (Figure 4a). Plotting the complex potential of a neuron accumulating these spiking inputs shows that it integrates the current caused by this stimulus into a voltage, exciting the neuron into a state which resonates with the succeeding input spikes (Figure 4b,d). Even if the input stops after a certain point in time, the neuron continues to produce spikes representing this input phase until its potential decays away through leakage and it ceases firing (Figure 4e).

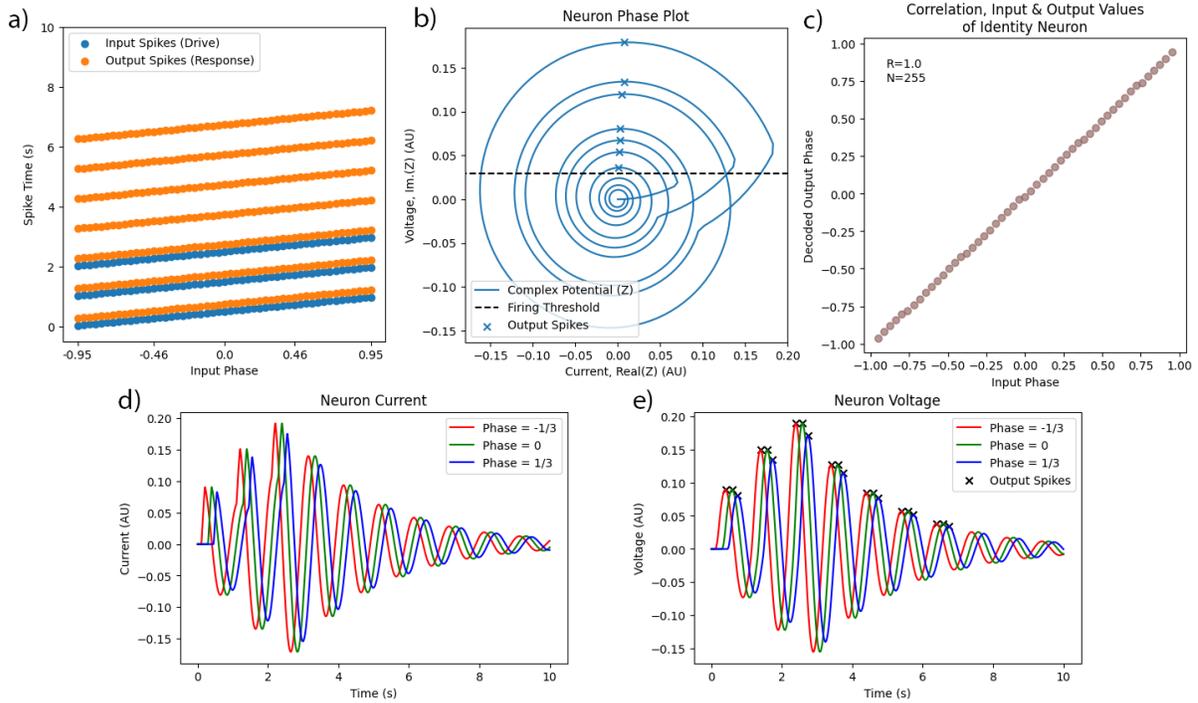

**Figure 4.**
(a) Regular input spikes with different relative phases (orange) cause a phasor neuron to resonate and produce a series of output spikes in turn (blue). (b) The resonance of a single neuron with a set of 3 input spikes is shown in a phase portrait. These input spikes cause jumps in current, exciting the neuron and causing its potential to oscillate. The peak of each voltage oscillation above the threshold produces a spike. This spiking continues even when the impulses cease, until the neuron's potential decays below the threshold value. (c) The phase of each neuron's 7 output spikes can be decoded using the initial starting time plus an integration delay (0.25 s). These decoded phases match the input phases to a high degree of precision. The difference in dynamics created by relative spike arrival times can be clearly seen here by comparing the currents (d) and voltages (e) of three phasor neurons driven by different inputs.

Each neuron's integration of current into voltage produces a quarter-period of delay ($T/4$) relative to its input. Adding this delay relative to the absolute time reference of the network's starting point, the phases of the output spikes can be decoded and compared to the ideal values encoded within the input spike trains. These decoded output phases successfully reproduce the values encoded in the input (Figure 4c; note, all the values belong to diagonal). This result shows that the complex summation and phase-detection can succeed at integrating and replicating a single input. However, the superposition of multiple inputs being calculated correctly through time is not addressed here, and is evaluated in the next section.

*Demonstration of Equivalent Layer*

To demonstrate that beyond the identity function, a temporal phasor neuron can calculate an approximately correct weighted superposition of its inputs (as described in Equation 11), we create a series of neurons corresponding to a 'layer' in a conventional network. The input weights to this temporal layer are taken from the hidden layer of an atemporal phasor network after it was trained on the standard MNIST dataset. A series of stimuli representing random input phases are applied to each neuron, which produce a series of output responses (Figure 5a). The conversions between spikes and phases at each layer inputs and outputs are identical to what we previously described. However, each neuron is now a subject to spikes from multiple (784) sources and it has to accurately integrate the weighted sum of these inputs to produce a correctly-timed output spike.

Importantly, the approximations used in Equation 11 (i.e., introduction of leakage and box kernel) do not prevent a temporal phasor neuron from producing an output representing a phase which is highly correlated to its atemporal value. Successive cycles of integration allow the network to produce output spikes which more closely match the atemporal values (R=0.93 in the final integration period) (Figure 5b,c). This demonstrates that a temporally-executed layer of phasor neurons can produce a good approximation of its atemporal counterpart.

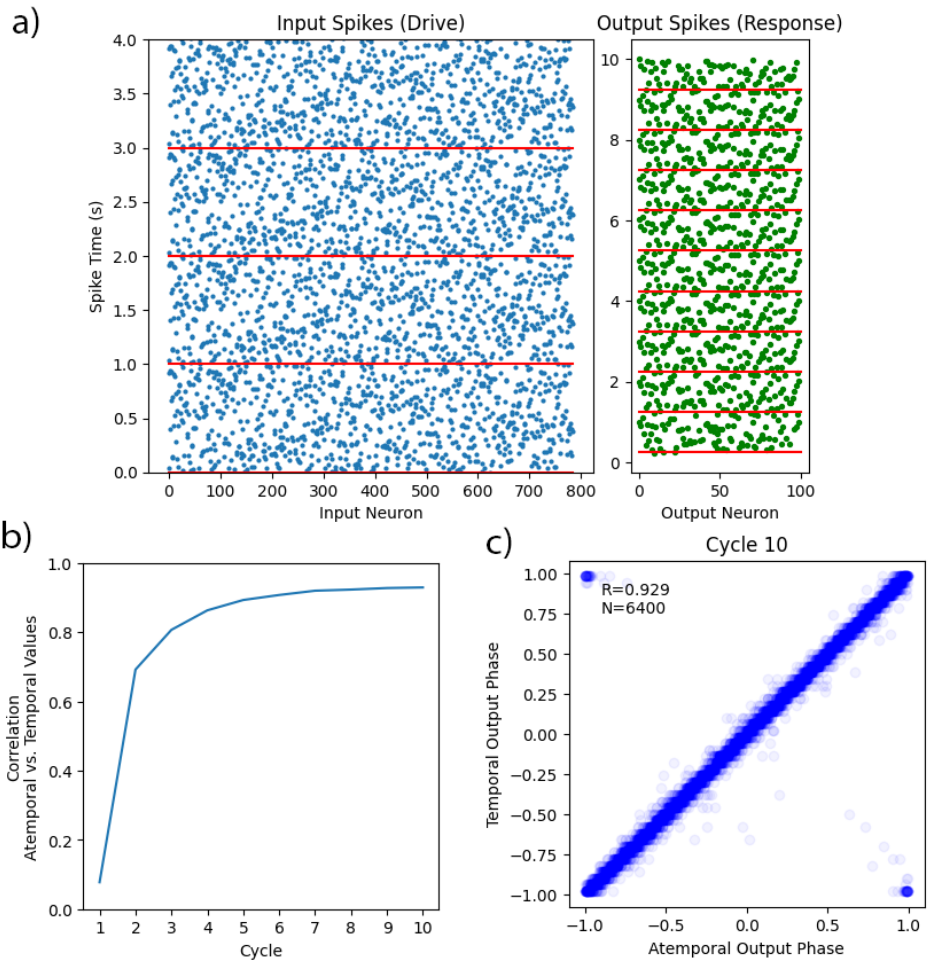

**Figure 5.**
(a) A spike raster shows one example of a spiking input which stimulates a layer of phasor neurons that produces a series of spikes in response. Horizontal red lines demarcate the boundaries between integration periods. (b) After several periods of integration, the spike phases decoded from the temporal network using an absolute time reference and the ideal values produced by the atemporal network are highly correlated. (c) This correlation reaches its peak value during the last executed integration cycle.

*Demonstration of Equivalent Network*

Given the approximate equivalence of a single phasor layer executed via spikes in the temporal domain to its atemporal execution mode (Figure 5), we next tested performance of the full networks used for image classification tasks. The networks were created identically to the networks demonstrated in the experiments above: both the MLP and convolutional networks were trained in the atemporal domain using standard backpropagation to reduce the loss function described in Equation 5. However, to test performance of the temporal model execution, instead of executing a trained network by passing tensors of values from layer to layer (atemporal execution), here the networks execute by sending precisely-timed binary spikes between layers (temporal execution). Network parameters remain identical between execution modes.

Inputs to the network executed in temporal domain are provided by stimulating the input layer with a series of spikes encoding the phase of input into their relative timing as previously described. These input impulses repeat every cycle during the execution of the network (Figure 6b). The predicted output class of an image is decoded from the output spike train by detecting the phase of spikes produced during the output layer's last full execution cycle (i.e. the final full cycle before the simulation's stopping time is reached). The phase is decoded by referencing to the initial start time, and the predicted label is defined by the neuron with the phase closest to the target value of ½ (Figure 6d).

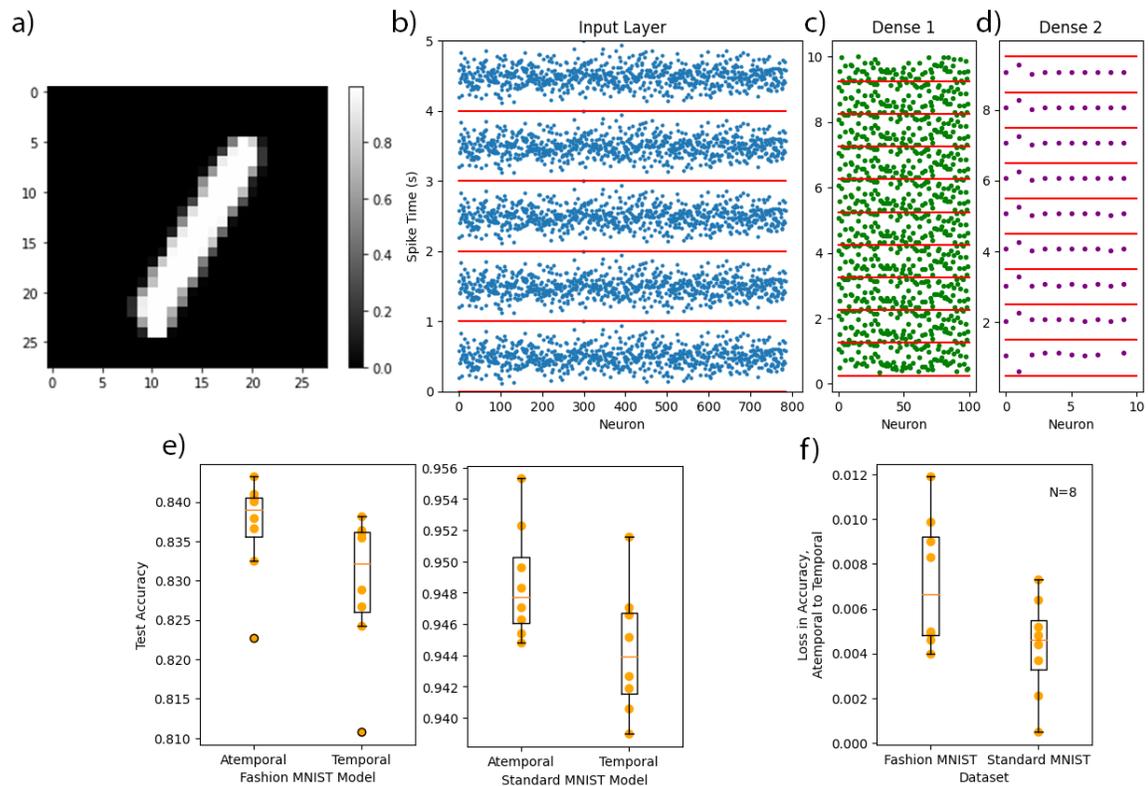

**Figure 6.**
(a) An input image from the MNIST dataset of the digit '1'. This image is (1) converted from intensities to phases and (2) into a spike train (b) which drives the phasor network in the temporal domain (c). Horizontal red lines demarcate the boundaries between integration periods. The network has been trained to produce an out-of-phase spike on the image label ('1', d). (e) 2 sets of 8 networks trained on the MNIST and FashionMNIST datasets are evaluated in both their atemporal and temporal execution methods, and the resulting accuracies on the test set are reported. (f) Running in temporal evaluation mode, most models lost an average of only 0.57% accuracy compared to atemporal evaluation.

We find that despite the significant change in underlying execution strategy – from standard matrix multiplication and activation function to the integration of spike-driven currents and phase detection through time – the final accuracy of MLP networks differs little between execution modes (Figure 6f). After an input image is flattened and converted to spike trains, each spiking layer of the network can perform an integration through time with sufficiently high accuracy to produce the desired output (Figure 6a-d). Results from 2 sets of 8 networks trained on the standard MNIST or F-MNIST datasets show that the two execution modes produce similar results and, on average, only 0.57% accuracy is lost by switching from atemporal to temporal execution (Figure 6e-f).

Next, a phasor-based convolutional network was executed temporally to test consequences of having a wider and deeper network architecture when using the temporal execution method. To temporally execute the min-pool operation, spikes within the pooling groups were examined cycle-to-cycle, and the earliest spike time in the pool was selected to be output. Due to greater resources required to simulate the temporal execution of this network with 70,514 neurons (versus 110 for the previous networks), a smaller test set of 1024 images was used. Running a single network on this reduced CIFAR test set, the network reached accuracies of 71.4% and 73.5% for temporal and atemporal execution modes, respectively. This corresponds to 2.1% accuracy loss by switching to the temporal execution mode.

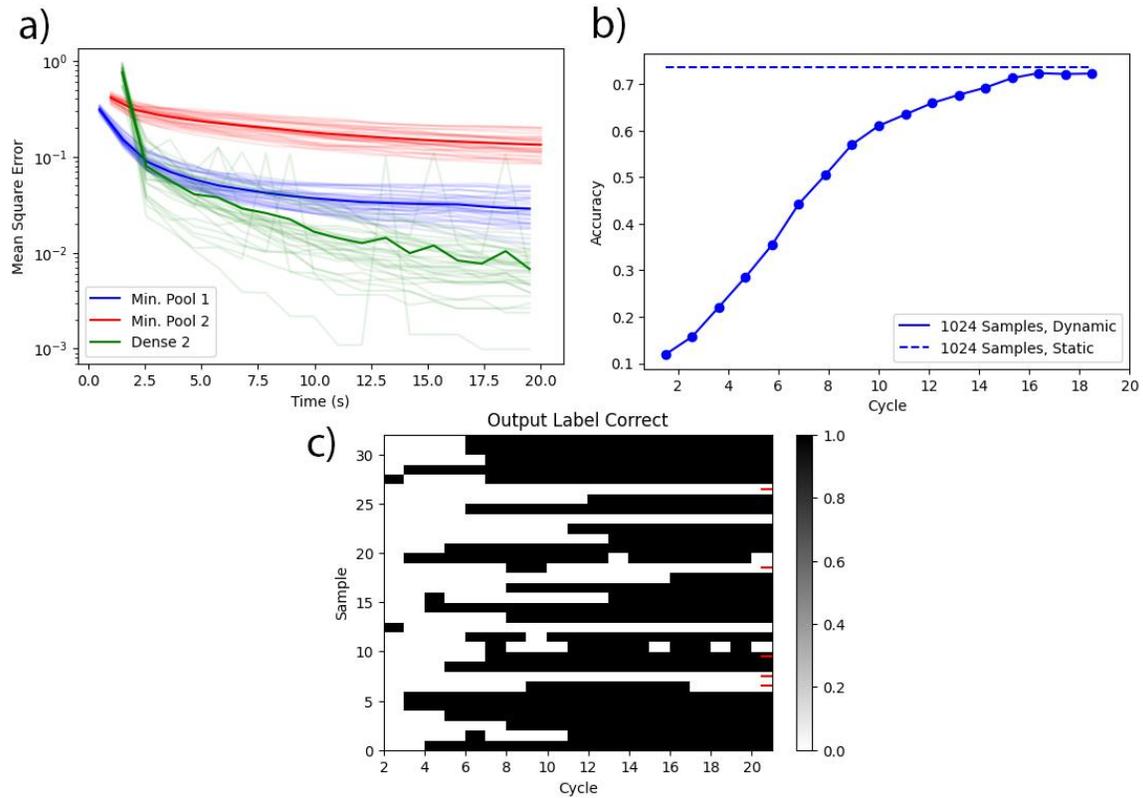

**Figure 7.**
(a) The mean squared error of phases produced by the temporal CIFAR network is computed by decoding the phases represented by temporal spikes and comping these values to those produced by atemporal execution. Results are presented from the outputs of the convolutional and dense blocks. Information propagates quickly from layer to layer, initially decaying more than exponentially. (b) The accuracy of the labels decoded at the spiking output with respect to time is plotted against the static accuracy for the given sample set. Within 15 execution cycles, the CIFAR network can reach an accuracy comparable to its static counterpart. (c) The correctness of decoded output labels with respect to time is shown, with each image corresponding to a row. Some images quickly reach the correct output, and others require more time for error within the network to be reduced. As expected, images misclassified under atemporal execution (marked with red ticks on the right side) are more likely to be misclassified under dynamic execution as well.

The greater depth and width of the network used for CIFAR data set allows for a more in-depth investigation of the propagation of information through layers with respect to time. This was measured by calculating the mean squared error (MSE) between the phases encoded by spikes in the network using temporal mode and the ideal phases calculated in the network under atemporal execution. We find that the initial execution cycles show the largest deceases in MSE, which continue to decay in time as more spikes propagate forward (Figure 7a). Different layers reach different lower bounds in error, and these differences stem from the approximate integration carried out by the resonate-and-fire neuron. A low error (approximately 1% MSE) is required at the dense output layer in order to reach temporal mode network classification performance comparable to atemporal execution mode (Figure 7b). As the temporal execution network approximates the same calculations carried out under atemporal execution, misclassified images are usually shared between execution modes (Figure 7c).

*Efficiency & Sensitivity Analysis*
There exist a variety of methods which allow for the execution of deep neural networks via sparse, spike-based dynamics. Each method is often related to the encoding of information in spikes via a certain coding scheme (33). Methods utilizing rate-based and time-to-first-spike (TTFS) coding have previously been successful in allowing spiking neural networks to achieve high accuracies on image classification tasks (9,18). However, each method carries a set of trade-offs between factors such as accuracy, efficiency, robustness, and more.

Rate-coding requires moderate integration periods in order for each neuron in the network to accumulate sufficient spikes from the previous layer to begin producing an accurate output in turn (9). This can limit the propagation of information from layer to layer, and different inputs can produce different spiking statistics (e.g. brighter images with higher initial activations may require more spikes). High spiking rates may lead to issues such as routing congestion on a neuromorphic chip (21). However, the redundancy of information within a rate-coded network can provide robustness of execution. Perturbation of an individual spike's timing is not highly detrimental, as only the long-term accumulation of spikes is used as the basis of computation.

In contrast, temporal codes use the timing of spikes relative to a local or global reference to communicate values. Time-to-first-spike (TTFS) is a common temporal code, in which the duration between a global starting point and the arrival of a single spike is used to encode values. This can greatly reduce the number of spikes required for a spiking network inference compared to a rate-coded equivalent (18). However, the sparsity enforced by this scheme gives it a higher vulnerability to perturbations such as the loss of spikes or disruptions in their timing; in contrast to rate-coding, the delay or loss of a single spike will communicate a different value.

We posit that phasor networks can strike a balance between the robustness provided by rate-coding and the efficiency provided by TTFS. The coding scheme used in phasor networks enforces one spike per integration period, but multiple integration periods are used per inference (Figure 6b). The 'momentum' of a neuron's potential from one cycle to the next can, over time, allow it to compute accurate outputs despite perturbations such as the loss of spikes between hidden layers (Figure 8a) or slight jitter added to the timing of spikes (Figure 8b). Additionally, one potential challenge of temporal codes is that they may require a higher level of time discretization used in their execution to be effective. We find that this is not the case for phasor networks; in the networks tested, approximately 40 points per integration cycle were found to be sufficient to execute both the MLP and convolutional networks (Figure 8c).

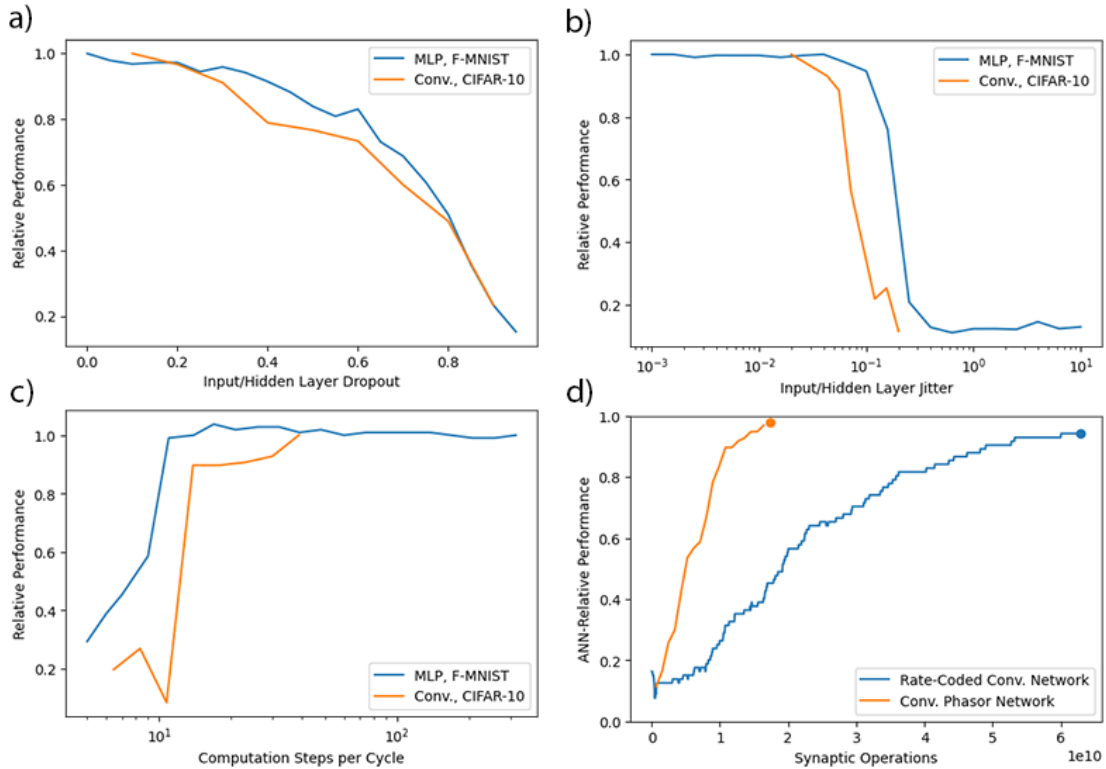

**Figure 8.**

The effect of various perturbations on a network's performance is measured by comparing the performance of the perturbed network in comparison to its original value. (a) By randomly removing spikes transmitted between layers, the sensitivity of phasor networks to loss of information between neurons is estimated. Networks remain robust to a moderate amount of dropout, and higher levels of dropout begin to affect performance more severely. These higher levels of dropout may not provide many neurons with sufficient drive to fire, leading to a larger loss in accuracy. (b) Randomly perturbing the timing of spikes produced by each layer estimates the sensitivity of the network to disruption of a temporal nature, such as random delays. Networks can tolerate a small amount of jitter, with the deeper convolutional network exhibiting a higher sensitivity to disruptions. (c) Related to timing jitter is the precision of the underlying integration being carried out. By varying the number of points per integration cycle used to calculate a solution, we find that 40 or more steps per cycle appears to be sufficient. (d) Synaptic operations provide a basis to estimate the computational resources required for a task (in this case, a classification inference). In comparison to a rate-coded network, a convolutional phasor network requires 29% of the synaptic operations to reach high accuracy.

While these points demonstrate the robustness of phasor networks against perturbations, it does not demonstrate their efficiencies relative to rate-coding or other temporal codes. By utilizing open-source software provided by Rueckauer et al., we execute the 'standard' version of our convolutional network (Figure 2a) via rate-coding (9). The spikes produced by each neuron in conjunction its fanout is used to compute the number of synaptic operations required for each network in comparison to the accuracy of its current predictions. Our results suggest that phasor networks do indeed provide a 'middle point' between rate-coding and TTFS; to reach a high level of accuracy compared to a non-spiking version, a phasor networks requires approximately 3.4x fewer synaptic operations than a rate-code based equivalent (Figure 8d). This is lesser than the 7-10x reduction reported for TTFS encoding (18), but we believe that the efficiency of phasor networks may be further improved by incorporating sparsity schemes. This point is elaborated in greater detail in the discussion below.

**Discussion**

In this study, we showed that the real valued activation function of a standard artificial neural network can be replaced by one that represents the angle of a complex value, or 'phasor'. With this approach, each neuron in the network integrates the phases of a set of inputs and generates a new output phase, which then propagates to the next layers. If all inputs are known ahead of time and with high precision, this calculation can be done in a time-agnostic or 'atemporal' manner. However, if all inputs are not known ahead of time, the calculation can also be carried out in real-time, integrating the signals as they arrive to produce a dynamic, time-varying calculation. In the former case, values are communicated via tensors of phase values, and in the latter case, communications are carried out using spikes. We believe that this model brings deep learning closer to biological relevance while maintaining key advantages over other spike-based deep learning models.

*Execution*

One of the important advantages of phasor representation is its ability to present the network with a complete input within a set time interval defined by the neuron's resonant frequency; each phase is encoded by the spike timing (its offset within the defined interval) and not by the accumulated rate of spikes (as is the case in rate-coding). This allows the network to produce outputs on an established time basis, rather than only after waiting an arbitrarily defined amount of time that is needed to estimate spike rate (as shown in Figure 6). Additionally, the dynamic coupling between layers transmits information through the network with only small delays, although time is required for the temporal outputs to reach a high degree of precision (Figure 7). The combination of these features allows phasor networks to more efficiently use synaptic operations in comparison to rate-coded networks (Figure 8d). In future works we aim to verify this in a greater number of architectures such as modern, very deep networks with real-world applications such as image detectors (34) and on real neuromorphic hardware.

An argument which can be presented against temporal phasor networks is that they may trade off the long integration times of rate-based networks by instead using a continuous time domain which may consist of an equivalent or even greater number of discrete steps when calculated on digital, clocked hardware. However, we do not find this to be an issue as phasor networks can be executed with approximately only 40 evaluation points per integration period without degrading performance (Figure 8c).

The oscillatory dynamics inherent in the temporal processing of a phasor neuron offers the possibility of a physical implementation of each neuron via a variety of novel analog hardware devices. The field of oscillatory computing has explored many such devices which are specifically designed to carry out computations via coupled oscillators (35). These devices can exploit the natural dynamics of electrical or mechanical oscillations to compute with extremely high efficiency. Combining these two approaches could therefore offer a potential application for new emerging systems by providing a framework for executing traditional AI methods with high efficiency via coupled oscillations.

Additionally, phasor networks are highly suited to photonic implementations. Many photonic approaches to accelerating AI tasks rely on encoding values via the intensity of light which is modulated through devices such as Mach-Zender modulators (MZMs) (36–38). Phasor networks instead allow the relative phase of coherent radiation traveling through a photonic system to be leveraged as a computational variable (36). The activation function required by a phasor neuron can be accomplished via a combination of phase-shifters, branches, attenuators, and a gain medium such as a lasing cavity (39). This strategy potentially provides an alternative method of implementing an on-chip nonlinearity which has proven to be a challenge for photonic approaches towards AI (38) while allowing for the scale-up provided by methods such as wavelength division multiplexing (40).

It may also be possible to implement phasor networks on several current neuromorphic platforms; however, many of these platforms use integrate-and-fire dynamics, rather than resonate-and-fire or more complex models (5,41,42). However, previous work by Frady et al. (23) has demonstrated how resonate-and-fire networks can be implemented via network dynamics based on integrate-and-fire neurons, providing a possible method to extend phasor networks to execute via simpler neuronal dynamics.

*Training & Architecture*

Another challenge of neuromorphic, spike-based architectures is the problem of how they can be efficiently trained *in-situ*, rather than exist solely as inference-only conversions of conventional networks trained via standard backpropagation on traditional hardware. This complex challenge consists of many issues, both practical and theoretical. Two of the foremost practical issues include backpropagating errors through binary spikes and the high memory requirements of training methods (such as backpropagation through time (BPTT)) which address the recurrent, temporal nature of spiking networks (43,44).

The use of surrogate gradients to produce eligibility traces on synapses allows synaptic variables to be included as a part of a dynamical system solved forward through time, allowing an observer in the future to approximately infer how recent spikes contributed to the current neuronal state and assign error credit (43,45). But rather than using traditional BPTT to solve credit assignment with a phasor network, its continuous update may instead allow for adjoint sensitivity methods to solve the coupled system backwards in time for a short period (46). This would greatly reduce the memory requirements of assigning credit through time, potentially creating an efficient method for training phasor networks inherently in the temporal domain. Future work could also investigate the application of synaptic plasticity, local learning rules, and feedback alignment in phasor networks to avoid the larger problems of gradient and weight transport (43,47). Finally, being able to execute natively in the spiking domain, phasor networks may allow for native implementation of different off-line processing modes, including those dependent on local learning rules, to improve generalization, transfer of knowledge and reduce forgetting during sequential task training (48,49)

Lastly, the fundamental representation of information via phase values in this network provides a rich basis of computation which can allow future phasor networks to compute with radically different architectures than traditional networks. This is due to the fact that a vector of phase values as used in the networks is identical to the basis of information in a vector-symbolic architecture (VSA), the Fourier Holographic Reduced Representation (FHRR) (50,51). This enables vectors produced by each layer of a phasor network to be manipulated through vector-symbolic manipulations such as binding and bundling, allowing complex data structures to be built within the framework of the network's evaluation (52). Sparse versions of this system and its related operations have proven effective are also under investigation (23,53). Leveraging these sparse representations within future phasor networks would reduce the number of spikes necessary at certain layers, potentially allowing these networks to operate with even higher efficiency in comparison to rate-coded equivalents. These VSA operations can be used to build new generation computation architectures with capabilities such as the factorization of components into symbols, linking them to the emerging field of neurosymbolic computation while maintaining spike-compatible computation (54–57).

## Conclusion

We demonstrated that by replacing the activation function of a standard feed-forward network by one created via complex operations, a novel 'phasor' network may be designed that can be executed either temporally or atemporally with no conversion process and only slight differences in output. The spiking and oscillatory computations of the temporal phasor network have strong parallels to biological computation, can be adapted to current or future neuromorphic hardware, and provide a rich basis for the construction of novel training methods and architectures.

## Materials and Methods

*Code Availability*
Networks were created and executed under Python 3.8.5 with Tensorflow 2.4.1 and SciPy 1.4.0. The full environment and all code/parameters necessary to generate figures are included in a public GitHub repository (https://github.com/wilkieolin/phasor_networks).

*Proof of Equation 10*

$$\frac{\partial z}{\partial t} = i\omega z + \sum_{i=1}^{n} w_i \delta(t - x_i)$$

$$z(T) = \int_0^2 [i\omega z + \sum_{i=1}^{n} w_i \delta(t - x_i)] \partial t$$

Given the assumptions:
1. An initial condition $z(t = 0) = 0$.
2. The previous bounds $x \in [-1,1]$ are shifted to $x \in [0,2]$ by adding 2 to all elements of $x$ less than zero. This is equivalent to offsetting traditional radian-based values by $2\pi$, leading to an identical phase.
3. The elements of $x$ are unique and sorted in ascending order.

$$z(T) = \int_0^2 \sum_{i=1}^{n} w_i \delta(t - x_i) \partial t$$

The first term is removed given the initial condition and the integration is performed only over the summation of currents. The element $x_1$ will have the lowest value and thus be the first delta function to be 'activated' with respect to time.

$$z(x_1^+) = \int_0^{x_1^+} w_1 \delta(t - x_1) + \sum_{n=2}^{n} w_i \delta(t - x_i) \partial t$$

$$z(x_1^+) = w_1 \cdot 1 = w_1 e^{i\pi 0}$$

This contribution in the real domain represents the 'current' from synapse $i$ which arrives at phase $x$. The sinusoidal oscillation between current and voltage in the R&F neuron is handled through the first term of the update equation which now has a non-zero value.

$$z(x_2^+) = \int_{x_1^+}^{x_2^+} i\omega z(x_1^+) + \sum_{n=2}^{n} w_i \delta(t - x_i) \partial t$$

$$z(x_2^+) = w_1 e^{i\pi(x_2 - x_1)} + w_2 e^{i\pi 0}$$

By induction, the following sum at t=2 is produced:

$$z(2) = \sum_{i=1}^{n} w_i e^{i\pi(2-x_i)} = \sum_{i=1}^{n} w_i e^{-i\pi x_i}$$

**Acknowledgments**

This work was supported by NIH T-32 Training Grant (5T32MH020002), the Lifelong Learning Machines program from DARPA/MTO (HR0011-18-2-0021) and ONR (MURI: N00014-16-1-2829).

We would like to thank Friedrich Sommer, E. Paxon Frady, Stefan Preble, and Matthew van Niekerk for their comments and discussions related to this work.